\documentclass[runningheads, mobile]{llncs}

\usepackage{eccv}

\usepackage{eccvabbrv}

\usepackage{graphicx}
\usepackage{booktabs}
\usepackage{cite}
\usepackage{amsmath,amssymb,amsfonts}
\usepackage{algorithmic}
\usepackage{graphicx}
\usepackage{multicol}
\usepackage{url} 
\usepackage{tabularx}
\usepackage{hyperref}
\usepackage{ltablex, multirow, multicol, makecell, caption}
\usepackage{bm}

\usepackage{textcomp}
\usepackage{xcolor}
\def\BibTeX{{\rm B\kern-.05em{\sc i\kern-.025em b}\kern-.08em
    T\kern-.1667em\lower.7ex\hbox{E}\kern-.125emX}}

\usepackage[accsupp]{axessibility}  

\usepackage{hyperref}

\usepackage{orcidlink}

\begin{document}

\title{Monitoring Viewer Attention During Online Ads} 


\author{Mina Bishay \and
Graham Page \and
Waleed Emad \and
Mohammad Mavadati}

\authorrunning{M.~Bishay et al.}

\institute{Affectiva, A Smart Eye Company \\
\email{\{mina.bishay, graham.page, waleed.emad, mohammad.mavadati\}@smarteye.ai}}

\maketitle

\begin{abstract}
Nowadays, video ads spread through numerous online platforms, and are being watched by millions of viewers worldwide. Big brands gauge the liking and purchase intent of their new ads, by analyzing the facial responses of viewers recruited online to watch the ads from home or work.  Although this approach captures naturalistic responses, it is susceptible to distractions inherent in the participants' environments, such as a movie playing on TV, a colleague speaking, or mobile notifications. Inattentive participants should get flagged and eliminated to avoid skewing the ad-testing process. In this paper we introduce an architecture for monitoring viewer attention during online ads. Leveraging two behavior analysis toolkits; AFFDEX 2.0 and SmartEye SDK, we extract low-level facial features encompassing facial expressions, head pose, and gaze direction. These features are then combined to extract high-level features that include estimated gaze on the screen plane, yawning, speaking, etc — this enables the identification of four primary distractors; off-screen gaze, drowsiness, speaking, and unattended screen. Our architecture tailors the gaze settings according to the device type (desktop or mobile). We validate our architecture first on datasets annotated for specific distractors, and then on a real-world ad testing dataset with various distractors. The proposed architecture shows promising results in detecting distraction across both desktop and mobile devices.


\end{abstract}

\section{Introduction}
\label{intro}

The global expenditure on online advertising continues to rise each year, reaching approximately \$602 billion in 2023 \cite{link_1}. Leading brands collaborate with international marketing agencies to assess the effectiveness of their new ads. In this process, individuals are recruited through an online platform to watch the ads and respond to related survey questions \cite{mcduff2016applications}. Participants provide consent to have their facial responses recorded as they watch the ads in their homes or workplaces, using either a desktop or mobile device. Then, the facial expressions are monitored, analyzed, and combined with survey responses to gain valuable insights about customers' level of engagement, liking, and purchase intent \cite{mcduff2013predicting, mcduff2014predicting}. Ensuring that participants remain attentive during the ad viewing is crucial for obtaining accurate insights from the ad-testing process. 


One of the primary distractors encountered in ad testing is when the participants divert their gaze from the screen to engage in other activities, such as texting, watching TV, or speaking. Gazing off the screen can be described by two behaviors; \textbf{owl behavior}, characterized by moving the head when gazing away, and \textbf{lizard behavior}, which involves primarily moving the eyes \cite{victor2015analysis, lee2016investigating}. Detecting these behaviors accurately necessitates a combination of head and eye tracking technologies. Other distractors include speaking which occurs frequently in environments with multiple people, drowsiness happening during lengthy ad content, and leaving the screen unattended. These distractors divert participants' focus from the ad, and should therefore be tracked and eliminated from any subsequent facial expression analysis.


Limited research has delved into monitoring attention during online ads \cite{mcduff2016affdex, schulc2019automatic}. While these studies focused on estimating head pose or gaze direction to identify instances of diverted gaze, they disregard critical  parameters such as device type (desktop or mobile), camera placement relative to the screen, and screen size. These factors significantly influence attention detection. In this paper, we propose an architecture for attention detection that encompasses detecting various distractors, including both the owl and lizard behavior of gazing off-screen, speaking, drowsiness (through yawning and prolonged eye closure), and leaving screen unattended. Unlike previous approaches, our method integrates device-specific features such as device type, camera placement, screen size (for desktops), and camera orientation (for mobile devices) with the raw gaze estimation to enhance attention detection accuracy.

The proposed architecture leverages two facial behaviour analysis toolkits; AFFDEX 2.0 \cite{bishay2023affdex} and SmartEye SDK \cite{link_2}, to detect four key distractors: gazing off-screen, speaking, drowsiness, and leaving the screen unattended. Using these toolkits we extract a range of low-level features tailored to each distractor. For instance, head pose and gaze direction are used to identify diverted gaze; mouth landmarks detect speaking; AUs and mouth aspect ratio identify yawning; and face detection/existence indicates unattended screens. These low-level features are then processed using trained models to derive high-level features, such as estimated gaze on the screen plane, screen size, speaking and yawning. These high-level features are capable of identifying the four distractors.


To assess the effectiveness of our architecture, we evaluate its performance across four distinct datasets. The first dataset involves participants who were instructed to direct their gaze at different screen locations, as well as off the screen at four different directions. The second and the third datasets contains videos captured in real-world scenarios, that are annotated for speech activity and yawning (active or inactive). Lastly, the fourth dataset consists of videos randomly selected from a real ad-testing dataset with various distractors. Experimental results demonstrate the efficacy of our proposed method in detecting key distractors (i.e. attention) across both desktop and mobile devices. 


The rest of the paper is organised as follows: we first review the related literature in attention detection in Section 2. We introduce the datasets used in our analysis in Section 3. Then, we describe the proposed architecture in Section 4. Finally, we give the experimental results and conclusion in Section 5 and Section 6, respectively.




\section{Literature Review}


Monitoring attention or distraction levels of individuals engaged in activities such as driving, attending lectures, or watching online content can provide valuable insights to enhance the outcomes of these activities. The work in attention detection can be broadly categorised into three primary applications; a) enhancing vehicle safety by identifying driver distraction, b) assessing student attention during online or in-person lectures, and c) monitoring viewer attention while watching online ads. In this section, we will review some of the works published within each of these categories, focusing solely on those leveraging Computer Vision techniques. 


\textbf{Driver distraction} accounts for 25\% of car accidents in the USA \cite{national2013preliminary}, underscoring the critical need to monitor and alert drivers to potential distractions, thereby potentially saving lives on the road. In recent years, the research community has shown significant interest in detecting driver distraction \cite{fernandez2016driver, kashevnik2021driver, misra2023detection}. Most of the existing research has focused on identifying three types of distraction: biomechanical, visual, and cognitive. \textit{Biomechanical distraction} involves removing hands from the steering wheel for doing a hand-related activity, such as texting, eating, adjusting makeup. In this category, the proposed approaches have primarily focused on either tracking the driver's hands \cite{ohn2014head, roy2022robust}, or detecting the hand-related activities \cite{xing2019driver, aboah2023deepsegmenter}. \textit{Visual distraction} occurs when drivers divert their eyes from the road, typically detected by estimating gaze direction using head and eye pose \cite{boyraz2012computer, vicente2015driver}. Lastly, \textit{cognitive distraction} arises when drivers become mentally distracted by non-driving thoughts. Several studies in this category have been reviewed by \cite{misra2023detection}.


\textbf{Students' attention} is crucial for promoting the learning process. The COVID-19 pandemic and the advances in digitization have transformed the learning environment, extending it from traditional physical classrooms to online/digital platforms, and Virtual Reality (VR) environments. Numerous studies have been dedicated to monitoring student attentiveness across different environments, encompassing traditional classrooms \cite{ngoc2019computer, lin2021student, trabelsi2023real},  digital/online platforms \cite{sharma2022student, hossen2023attention}, and VR environments \cite{rahman2020exploring, asish2021deep, zarour2023distraction}. These methodologies typically involve monitoring eye, head, and body pose for detecting the different attentive (e.g. raising hand, gazing at board/screen) and inattentive (e.g. gazing away, laughing, using the phone) activities.

\textbf{Viewers' attention} towards ads is a critical aspect in digital advertising. However, there has been limited research focusing on detecting attention during online ads. In \cite{mcduff2016affdex}, attention was gauged solely based on head pose, where participants were deemed inattentive if their head angle exceeded a certain threshold. In \cite{schulc2019automatic}, a large dataset consisting of $\sim$28K participants was annotated for attention, where inattentive activities include gazing off-screen, closing eyes, or engaging in unrelated activities (e.g. telephoning). A CNN-LSTM model was trained for detecting attention based on temporal facial appearance.  




Previous studies in ad testing overlooked important factors such as device type (desktop/mobile), screen size, and camera orientation in their analyses, relying solely on head pose or eyes appearance as indicators of gazing off-screen. Moreover, \cite{mcduff2016affdex} focused only on identifying diverted gazes, and neglected other potential distractors. In contrast, \cite{schulc2019automatic} trained a single shallow CNN for detecting complex activities, including talking, gazing off-screen, eating, and drowsiness. However, this approach may have exceeded the network's learning capacity. 

It is important to note that detecting attention in ad testing involves different settings compared to the driving and educational scenarios. First, eye tracking in ad testing is conducted without any calibration or defined camera integration, whereas driving and educational sessions typically have well-defined calibration and integration processes. Second, ad testing focuses on desktop and mobile devices, requiring highly accurate eye trackers to function within the limited gaze range of these devices. In contrast, driving scenarios, which involve various car components like windshields, center consoles, and mirrors, as well as educational sessions that include whiteboards, projector screens, and desktop devices, call for moderately accurate trackers. This is because these settings demand a broader gaze range. Finally, ad testing monitors attention offline for quite short content (ranging from seconds to a few minutes), while in driving and educational scenarios, attention is monitored continuously in real time for extended periods.

\section{Datasets}

In this section, we describe four datasets used for training and testing our attention architecture. The first dataset focuses on gazing on/off the screen, the second on speaking, the third on yawning, and the fourth is a real ad-testing dataset with various distractors. We used behavior-specific datasets for gazing off-screen, speaking, and yawning because capturing a diverse set of training examples for these behaviors from a real dataset would require labeling a much larger number of videos. These datasets mainly come from our large ad-testing repository, built over years through collaborations with global market agencies. The repository consists of millions of sessions recorded for participants watching commercial ads worldwide (from over 100 countries). For all these sessions, a web-based approach is used to record participants in their homes or workplaces (i.e. in the wild), after getting their consent.



\subsection{Gazing Dataset}

We employ the same web-based approach to collect a controlled dataset for gazing on/off the screen. Participants were recruited to watch a video containing instructions prompting them to first follow a dot across different screen locations, including the far screen edges. Then, the participants were instructed to divert their gaze off the screen in four different directions (up, down, left, right), with this process repeated three times. The moving dot modality is employed for collecting attentive examples, while the looking-off screen modality for collecting inattentive examples, and subsequently collecting a diverse set of positive and negative samples for training and testing. The duration of the video stimuli was approximately 160 seconds.

\begin{figure}[!t]
  \centering{\includegraphics[width=0.99\linewidth]{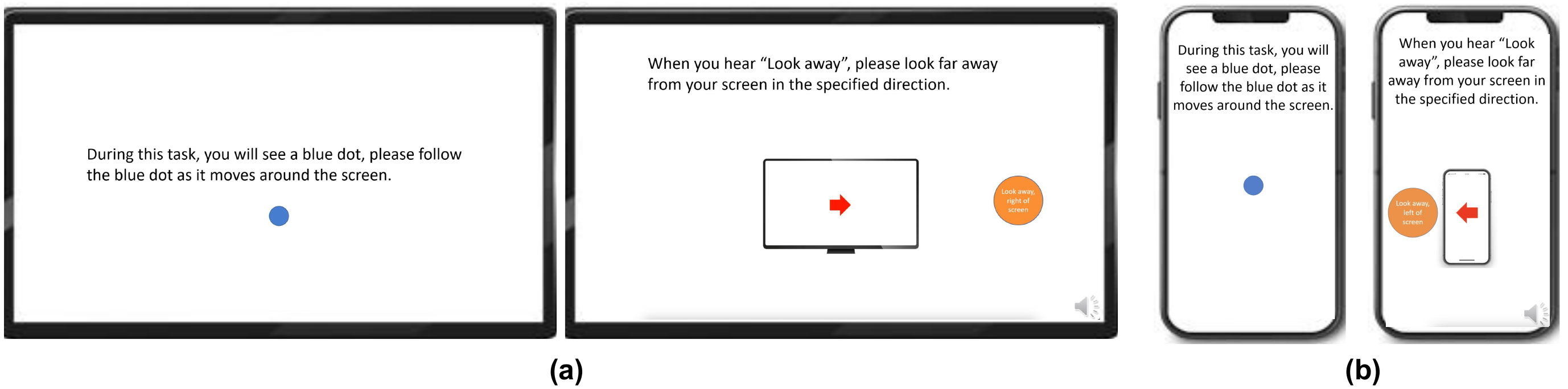}}
  \caption{Screenshots show the gaze video stimuli as presented on (a) desktop and (b) mobile devices. The first and third images display instructions given before moving the dot across the screen, while the second and fourth images show instructions asking participants to gaze off-screen. }
  \label{gaze_proj}
\end{figure}

Two versions of the video stimuli were designed: one for desktop devices with a resolution of 1920x1080, and another for mobile devices with a resolution of 608x1080. Desktop devices include laptops and standalone PCs, while mobile devices include tablets and cellphones. Examples of both versions are illustrated in Figure \ref{gaze_proj}. In total, we gathered 609 videos: 322 from desktop devices and 287 from mobile devices. The videos were automatically labeled based on the video stimuli: segments corresponding to instructions or the dot moving across screen were labeled as "gazing on-screen", while instances of looking away from the screen in any direction were labeled as "gazing off-screen". The dataset is divided into a training set of 158 videos and a testing set of 451 videos.

\subsection{Speaking Dataset}

Our goal is to detect speaking using visual cues alone, since audio signals are not captured during ad testing. To achieve this, we have collected and labeled a large dataset from our repository. This dataset is divided into two parts: the first part, which is manually labeled, is primarily used for training and validation, while the second part, which is automatically labeled, is dedicated for broader testing. The first part includes approximately 5.5K videos, each manually annotated by three labelers for speaking (speaking or no speaking). These videos are divided into 4.4K for training and validation, and 1.1K for testing. The second part of the dataset comprises a larger set of 16K recordings; 10.5K of participants watching ads with no expected speaking (non-conversational data) and 5.5K of participants expressing their opinions about different brands (conversational data). The 16K sessions are automatically annotated as either speaking or no speaking based on the data type. In the experimental section, we report the performance of our speaking model on both parts of the dataset.

\subsection{Yawning Dataset}

While a few datasets are publicly available for yawning \cite{fan2007yawning, abtahi2014yawdd}, none of them are suitable for detecting yawning in ad testing. The dataset by Fan \textit{et al.} \cite{fan2007yawning} includes only face images with yawning, which can be confounded with expressions of speaking, surprise, or fear, while the dataset by Abtahi \textit{et al.} \cite{abtahi2014yawdd} contains acted, non-spontaneous yawns captured in driving scenarios. Therefore, we created our own dataset for detecting yawning in the ad-testing scenario.

From our ad-testing repository, we selected 735 videos that potentially contain yawning events. These videos were chosen so as to have a high probability of jaw drop for more than a second. Each video was manually annotated by three labelers for yawning, either active or inactive. The ratio of frames with active yawning to the total number of frames in our dataset was 2.6\%, highlighting the challenging and imbalanced nature of the yawning dataset. The dataset is divided into a training set of 670 videos and a testing set of 65 videos.




\subsection{Real Distraction Dataset}

The real distraction dataset comes also from our large ad-testing repository, where people were recruited to watch real ads without any assigned tasks. This dataset displays different distractors, and will be used for evaluating the efficiency of our attention architecture. Specifically, we randomly select a total of 520 sessions, encompassing 193 mobile and 327 desktop sessions, from our repository. The dataset is manually annotated by 3 labellers for attention (attentive/inattentive). Inattentive events include gazing off-screen, speaking, drowsiness, and leaving screens unattended. The selected sessions are quite diverse in demographics as they have been collected across different regions; USA and Canada (34\%), Europe (25\%), Asia (28\%), Latin America (12\%), and others (1\%). Note that more sessions were included from desktop devices as webcam can be mounted at more varying locations compared to mobile devices that are typically mounted on top center. 

Unfortunately, the four datasets used in our analysis are proprietary data and cannot be made open source based on the participants' consent agreements.

\section{Attention Model}


\subsection{Overview}

\begin{figure*}[!t]
  \centering{\includegraphics[width=0.99\linewidth]{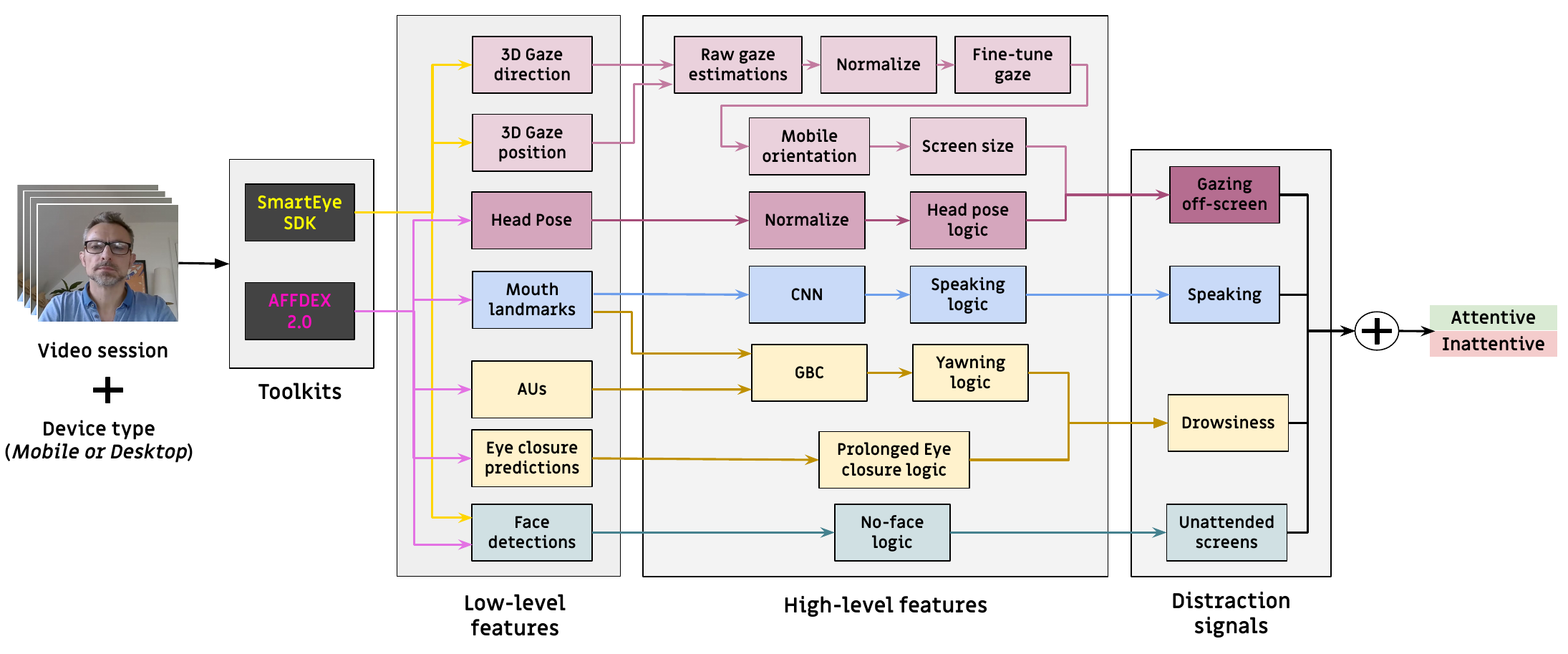}}
  \caption{The proposed architecture for monitoring attention during online ads. }
  \label{attention_framework}
\end{figure*}

In this section, we introduce our proposed method for monitoring viewers' attention during online ads. An overview of the proposed method is shown in Figure \ref{attention_framework}. Our model operates by two inputs: a video recorded for a participant watching an ad, and the type of the recording device (desktop or mobile). For every video frame, the model produces a binary label, indicating whether the participant was attentive or not. Attention is monitored by detecting four distraction behaviors often observed during ad watching; gazing off-screen, speaking, drowsiness, and unattended screens. Gazing off-screen is detected using either eye gaze or head pose, depending on the eye gaze quality.


Our model utilizes two facial analysis toolkits for extracting low-level features, that is, AFFDEX 2.0 \cite{bishay2023affdex} for detecting facial expressions/AUs and head pose, and the SmartEye SDK \cite{link_2} for tracking precisely the eye gaze. These features are combined to extract high-level features that are essential for identifying the four distractors. Both toolkits are designed to analyze the face in uncontrolled settings (i.e. in the wild), making them ideally suited for our application. In the following sections, we describe the modeling steps for detecting each distractor. Note that the gazing-away distractor is split into 2 parts/sections; one based on eye gaze and the other on head pose. 

%




\subsection{Gazing Off-Screen (Gaze Model)}

The direction of eye gaze serves as a powerful indicator of participant attention during ad watching. Specifically, participants are considered attentive if their gaze falls within the screen boundaries, and inattentive otherwise. Participants are typically recorded in home or work environments. To accurately detect attention in such uncontrolled scenarios, without calibration or defined camera integration, our gaze model should estimate the intersection of the gaze with the screen plane, along with several extrinsic parameters. These parameters include camera orientation for mobile devices, and screen size and camera location for desktop devices. The next subsections explain the methods employed for estimating the gaze-to-screen intersection and the extrinsic parameters. This approach effectively detects the \textit{owl behavior} and the \textit{lizard behavior} of gazing away.


\subsubsection{Gaze intersection with the screen plane} requires performing three processing steps: estimating the raw gaze coordinates on the screen plane, normalizing the estimated gaze across the whole video, and fine-tuning the normalized gaze using a trained model. 

\textbf{a. Raw gaze estimations.} We use the SmartEye SDK to track participants' eye gaze, which involves estimating the 3D gaze direction $\vec{\bm{D}}$, the 3D pupil position $\bm{p}=(x_{p}, y_{p}, z_{p})$, and the quality of the estimated gaze. These signals are essential for estimating the gaze intersection with the screen plane. Note that the screen plane is infinite and extends beyond the screen boundaries. According to analytic geometry, the screen plane $\bm{S}$ and the gaze ray $\bm{R}$ can be defined as:
\begin{equation}
S = x_{s}X + y_{s}Y + z_{s}Z, \\
\end{equation}
\begin{equation}
\begin{aligned}
R & = p + \vec{\bm{D}}t  \\  
  & = (x_{p}, y_{p}, z_{p}) + (x_{d}X + y_{d}Y + z_{d}Z)t, \\
\end{aligned}
\end{equation}
where $t \in \mathbb{R}$ is a scalar multiplied by the gaze vector. If we assume that the camera is positioned within the screen plane and that the screen plane is parallel to the participant's face, the equation for the plane can be simplified to:
\begin{equation}
S = x_{s}X + y_{s}Y.
\end{equation}

The point where the gaze ray intersects with the screen plane can be determined by equating the screen plane and the gaze ray equations:
\begin{equation}
(x_{p}, y_{p}, z_{p}) + (x_{d}X + y_{d}Y + z_{d}Z)t  =  x_{s}X + y_{s}Y,
\end{equation}
which means that the scalar $t$ can be calculated as: 
\begin{equation}
\begin{aligned}
z_{p} + z_{d}t = 0 \\ 
t = -z_{p} / z_{d}.
\end{aligned}
\end{equation}
The intersection point coordinates on the screen plane, $x_{s}$ and $y_{s}$ are equal to:
\begin{equation}
\begin{aligned}
x_{s} = x_{p} + x_{d}t \\
y_{s} = y_{p} + y_{d}t.
\end{aligned}
\end{equation}
$x_{s}$ and $y_{s}$ are compared to the screen boundaries, and if they fall within the boundaries then the participant is considered looking at the screen (i.e. attentive), and inattentive otherwise. 

\begin{figure*}[!t]
  \centering{\includegraphics[width=0.95\linewidth]{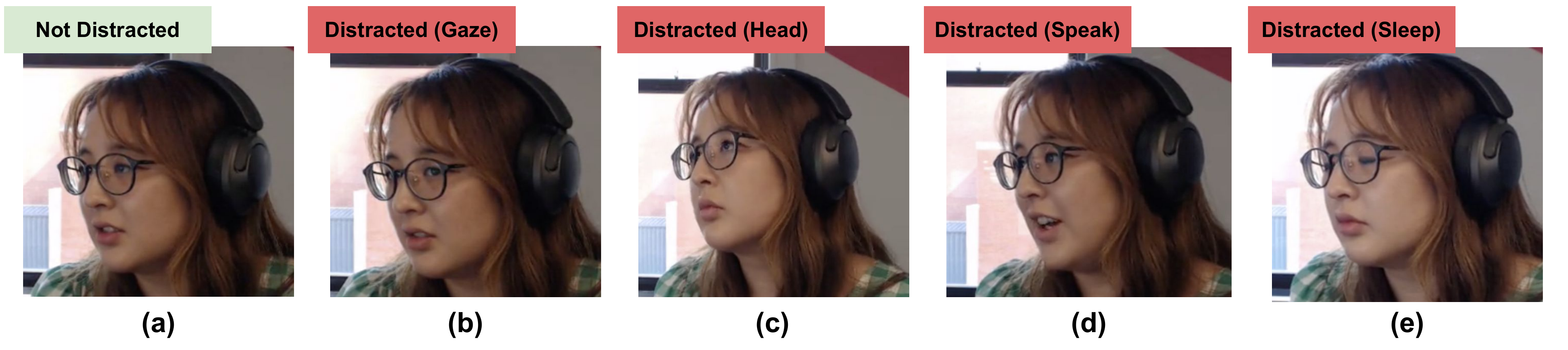}}
  \caption{Examples showing the true activation of the distraction signals on a video captured from a desktop device with webcam not centrally positioned. }
  \label{attention_results_non_frontal}
\end{figure*}

\textbf{b. Gaze normalization.} Although webcams are commonly integrated as built-in accessories, typically positioned at the top center of laptops and mobile devices, some desktop devices lack a built-in camera, leading users to position external webcams according to their preference. Additionally, some users have multiple screens and may mount the webcam on a different screen from the one displaying the ad. To mitigate the impact of multiple screens and varying camera locations, and to minimize the necessity for calibration, we normalize the raw gaze estimations by subtracting the mean gaze value across each recording. This normalization process is implemented for both mobile and desktop sessions. Figure \ref{attention_results_non_frontal} demonstrates how this normalization step allows our architecture to detect gaze distraction for participants using multiple screens.



\textbf{c. Fine-tuning the normalized gaze.} Assumptions, errors in estimations, and unaddressed parameters, such as focal length, distortion parameters, principal point deviate the gaze estimations slightly from their true locations. To reduce this deviation, we employ a Gradient Boosting Regressor \cite{friedman2001greedy, friedman2002stochastic} to minimize the error between the estimated gaze and ground truth. Specifically, the regressor takes as input the normalized screen coordinates ($x_{s}$ and $y_{s}$) and outputs more precise screen locations. The regressor is trained using the gazing dataset, that is, the gaze estimations corresponding to the dot moving across various screen locations, with the exact dot locations serving as the ground truth values. For training, 96 videos are utilized for desktop devices and 62 for mobile devices. Note that the gaze model is trained using only data within the screen, as ground truth labels cannot be generated for gazing off-screen.


\subsubsection{Determining extrinsic parameters} 
includes estimating the camera orientation for mobile devices, and the screen size/boundaries for desktop devices.

\textbf{a. Camera orientation.} The screen dimensions of mobile devices change with device rotation. To detect camera orientation, we first annotate 155 mobile sessions with one of 3 possible camera orientations; centered camera, rotated clockwise, or rotated anticlockwise. Then, we calculate three features for each session; average gaze across the $X$ dimension, average face location, and average head yaw angle. Finally, the 3 features are compared against predetermined thresholds, and combined using a majority voting approach to determine the camera orientation. 




\textbf{b. Screen size} is a crucial parameter in our model, especially given its significant variability across desktop devices. In \cite{sugawara2006future}, Sugawara \textit{et al.} highlight the relationship between the screen size and the preferred viewing distance. The distance between participants' eyes and the camera typically changes linearly with screen size. Therefore, we calculate the average eye-to-camera distance across the entire video and use the defined linear transformation in \cite{sugawara2006future} to determine the screen size.



\subsection{Gazing Off-Screen (Head Model)}

The accuracy of the gaze estimation model is influenced by the quality of the video recording, which can be affected by factors such as video resolution, illumination conditions, and head pose. While head pose tracking is generally more robust than the eye gaze tracking, we use the head pose as a complementary indicator for detecting instances of gazing off-screen, especially when the gaze estimation quality falls below a certain threshold. his method successfully identifies the \textit{owl behavior} of looking away.



The AFFDEX 2.0 toolkit is used to estimate the head pose, including yaw, pitch, and roll angles. The yaw angle corresponds to gazing left and right of the screen, while the pitch angle pertains to looking up and down the screen. In our model, we normalize the yaw and pitch angles by subtracting the mean across the entire video, similar to the normalization of the eye gaze. If the normalized pitch and yaw angles exceed predetermined thresholds (tuned according to the gazing dataset), the participant is considered inattentive, and attentive otherwise. 




\subsection{Speaking}

In ad testing, participants were recorded without capturing their audio signal. Consequently, we rely solely on visual cues to detect speaking by tracking mouth movements over time. To detect speaking, we first detect the dense landmarks around the mouth region using the AFFDEX 2.0 toolkit. Then, we calculate the vertical distance between the landmarks representing the upper and lower lips. The vertical distances across a temporal window of 1 second are concatenated and passed as input to a Convolutional Neural Network (CNN). The CNN consists of two 1-dimensional convolutional layers with kernels of width 3, and a fully-connected layer with a width of 50. The CNN is trained for 200 epochs.


During ad watching, speaking behavior can manifest as short events resulting from a highly engaging ad or longer events when participants are distracted by speaking with people nearby. To differentiate between these two behaviors, we use a 1-second time window. If speaking exceeds 1 second, it is considered inattentive behavior; otherwise, it is deemed an engaging response. 




\subsection{Drowsiness}

Our model detects drowsiness using two primary indicators: prolonged eye closure and yawning. The activation of either signal indicates drowsiness and suggests that the ad viewer is inattentive. To detect prolonged eye closure, we use the eye closure signal as a low-level indicator. We refine eye closure by excluding instances coinciding with smiling (AU12) or cheek raising (AU6), as this combination of AUs can occur in response to laughter or horror scenes (e.g., in horror trailers). AFFDEX 2.0 is used to detect eye closure, smiling, and cheek raising signals. Finally, we only consider eye closure events lasting longer than 2 seconds as inattentive moments, in order to avoid misidentifying blinks or sentimental moments as drowsy events.

Following the methods described in \cite{ji2004real, jie2018analysis, saurav2021real}, we detect yawning by calculating first the mouth aspect ratio (the ratio of the mouth height to width) using the mouth landmarks from AFFDEX 2.0. Next, we use the mouth aspect ratio along with the AU predictions to train a Gradient Boosting classifier (GBC) \cite{friedman2001greedy, friedman2002stochastic} for yawning detection.

\subsection{Unattended Screen}

To identify unattended screens or ads, we rely on the face detection signals provided by the AFFDEX 2.0 and SmartEye toolkits. In particular, if neither toolkit detects a face for a duration exceeding one second, we conclude that the ad is unattended and the participant is inattentive. 


The activation of the 5 distraction signals -- gazing off-screen (eye gaze), gazing off-screen (head pose), speaking, drowsiness, and unattended screen -- is combined together to form a single distraction/attention signal. Thus, our framework can identify the signals that contribute to the activation of the attention signal.





\section{Experimental Results}




In the following sections, we validate the various models used in our attention architecture -- gazing off-screen, speaking, yawning, and leaving screen unattended -- and assess the impact of combining these distraction signals for detecting attention.


\subsection{Validation of the Gaze Model}

The gaze model is designed to detect when a participant is gazing off-screen and involves several processing steps: a) normalizing the raw gaze estimations, b) fine-tuning the normalized gaze, and c) estimating the screen size for desktop devices. In this section, we evaluate the significance of each processing step by examining the impact of its removal from the gaze model. We use two datasets for evaluation; the real distraction dataset and the testing subset of the gazing dataset, comprising 226 desktop videos and 225 mobile videos. We report our model's performance using the Geometric Mean (G-mean) \cite{tharwat2021classification} and F1 scores. The G-mean aims to maximize the accuracy across both classes and is calculated as the geometric mean of the true positive rate and true negative rate.





Table~\ref{gaze_distr} illustrates the performance obtained by the full gaze model, as well as by individually removing each processing step. Across both datasets, it is evident that removing any processing step results in lower performance compared to the full gaze model. Furthermore, the normalization step has larger improvement on desktops than on mobile devices, as desktops are expected to have more varying camera locations than the standardized built-in cameras found in mobile devices.


\begin{table*}[!t]
\centering
\caption{Comparing the impact of removing the different processing steps from the gaze model.}
\label{gaze_distr}
\resizebox{0.99\textwidth}{!}{%
\begin{tabular}{ | c || c | c | c | c || c | c | c | c |} \hline

\multirow{3}{*}{\textbf{Gaze model}} & \multicolumn{4}{|c|}{\textbf{Real Distraction Dataset}} & \multicolumn{4}{|c|}{\textbf{Gazing Dataset}}  \\ \cline{2-9}

  & \multicolumn{2}{|c|}{\textbf{Desktop}} & \multicolumn{2}{|c|}{\textbf{Mobile}} & \multicolumn{2}{|c|}{\textbf{Desktop}} & \multicolumn{2}{|c|}{\textbf{Mobile}}  \\ \cline{2-9}

  & \textbf{G-mean} & \textbf{F1} &  \textbf{G-mean} & \textbf{F1}   & \textbf{G-mean} & \textbf{F1} &  \textbf{G-mean} & \textbf{F1}  \\ \hline  \hline

\textbf{w/o normalization}  & 0.678  & 0.330  & 0.795   & 0.560   & 0.656  & 0.523  & 0.740 & 0.595 \\ \hline

\textbf{w/o fine-tuning model}  & 0.639  & 0.443  & 0.802 & 0.651 & 0.659 & 0.557  & 0.751 & 0.645 \\ \hline


\textbf{w/o screen size detection (desktop only)}  & 0.726  & 0.479  & -  & - & 0.682  & 0.557  & - & - \\ \hline


\textbf{Eye gaze (full model)}  & \textbf{0.740}  & \textbf{0.504}  & \textbf{0.807} & \textbf{0.628}  & \textbf{0.685}  & \textbf{0.563}  & \textbf{0.761} & \textbf{0.647}  \\ \hline

\end{tabular}}
\end{table*}

Furthermore, we examine the effect of integrating various features -- face location, eye gaze, and head pose -- on detecting mobile camera orientation. We evaluate our model performance using the macro F1 score. The performance obtained for the face location, head pose, and eye gaze features are 0.75, 0.74, and 0.60, respectively. Combining all three features results in a notably improved performance of 0.91, which is significantly better than any single feature. 




\subsection{Validation of the Speaking Model}


In this section, we evaluate the performance of the trained speaking model using both the manually labeled testing set and the large automatically labeled dataset. We quantify the model's performance with the ROC-AUC metric. Our speaking model, trained using vertical lip distance, achieves a ROC-AUC of 0.97 on the testing set and 0.96 on the other large dataset. This demonstrates the robust performance of our speaking model across both evaluation phases.

\subsection{Validation of the Yawning Model}

In this section, we use the yawning testing set consisting of 65 videos for evaluating the trained yawning model. First, we evaluate the yawning model when using just the mouth aspect ratio. Then, we evaluate merging the mouth ratio with the 20 AU predictions from AFFDEX 2.0. Our model was able to achieve a ROC-AUC of 96.6\% when using the mouth ratio, and 97.5\% when adding the AUs to mouth ratio. This highlights how our model can effectively detect the yawning behaviour.


\subsection{Validation of the Unattended-Screen Model}

The unattended-screen model triggers when both the AFFDEX 2.0 and SmartEye toolkits are unable to detect a face for more than a second. Our model assumes that these no-face instances indicate unattended screens, representing moments of inattention. To validate this assumption, we manually annotate all the no-face activations in the real distraction dataset with the reason behind the activation. Then, we summarize the reasons for these activations and calculate the percentage for each reason. We exclude from our analysis activations that are unclear if they are indicating attentive or inattentive moments (e.g. covered camera, pixelated video).  

Table~\ref{no_face_table} presents the reasons for the no-face activation, and the occurring percentage for each reason. Additionally, we calculate the percentage of the no-face events representing truly inattentive moments (true activations) and attentive moments (false activations). Despite unattended screens constituted only 27\% of the instances triggering the no-face signal, it was activated for other reasons indicative of inattention, such as participants gazing off-screen with an extreme angle, doing excessive movement, or occluding their face significantly with an object/hand.


\begin{table}[!t]
\centering
\caption{The distribution of the different reasons related to the no-face signal.}
\label{no_face_table}
\resizebox{0.7\textwidth}{!}{%
\begin{tabular}{ | c | c | c |} \hline
\textbf{Reason}  &  \textbf{Reason}   & \textbf{Valid}   \\
\textbf{}  & \textbf{percentage}   & \textbf{activation (\%)?}  \\ \hline  \hline

Gazing off-screen w/ extreme angle & 31.5\%  &  \multirow{4}{*}{True (81\%)}   \\ \cline{1-2} 
Leaving screen unattended & 27.4\% &     \\ \cline{1-2} 
Face largely occluded by hand/object & 13.7\% &   \\ \cline{1-2} 
Excessive participant movement & 8.2\% &  \\ \hline

                          
Significantly cropped or darkened face & 16.4\%  & \multirow{2}{*}{False (19\%)}   \\ \cline{1-2} 
Rotated face & 2.7\% &   \\ \hline
\end{tabular}}
\end{table}
\setlength{\tabcolsep}{4pt}
\begin{table}[!t]
\begin{center}
\caption{Comparing the effect of adding the different distraction signals to our attention architecture.}
\label{various_signals}
\resizebox{\textwidth}{!}{%
\begin{tabular}{ | c || c | c | c | c || c | c | c | c |} \hline

\multirow{3}{*}{Model/Behaviour} & \multicolumn{4}{|c|}{\textbf{Real Distraction Dataset}} & \multicolumn{4}{|c|}{\textbf{Gazing Dataset}}  \\ \cline{2-9}

  & \multicolumn{2}{|c|}{\textbf{Desktop devices}} & \multicolumn{2}{|c|}{\textbf{Mobile devices}} & \multicolumn{2}{|c|}{\textbf{Desktop devices}} & \multicolumn{2}{|c|}{\textbf{Mobile devices}}  \\ \cline{2-9}

  & \textbf{G-mean} & \textbf{F1} &  \textbf{G-mean} & \textbf{F1}   & \textbf{G-mean} & \textbf{F1} &  \textbf{G-mean} & \textbf{F1}  \\ \hline  \hline

\textbf{AFFDEX 1.0 \cite{mcduff2016affdex}}  & 0.350  & 0.170  & 0.522   & 0.417   & 0.423  & 0.294  & 0.483 & 0.362 \\ \hline   \hline

\textbf{Gazing off-screen (head model)}  & 0.570  & 0.384 & 0.532  & 0.420  & 0.589  & 0.475 & 0.566 & 0.461  \\ \hline 


\textbf{+ Gazing off-screen (gaze model)}  & 0.788  & 0.531 & 0.797   & 0.650  & 0.706  & 0.588 & 0.756   & 0.650 \\ \hline
\textbf{+ Drowsiness}  & 0.789  & 0.533 & 0.802   & 0.655  & -  & - & -   & - \\ \hline
\textbf{+ Speaking}  & 0.804  & 0.548 & 0.834  & 0.687  & -  & - & -   & -    \\ \hline
\textbf{+ Unattended screen (all distractors)}  & \textbf{0.829}  & \textbf{0.585} & \textbf{0.876}   & \textbf{0.768} & \textbf{0.717}  & \textbf{0.601} & \textbf{0.773}   & \textbf{0.666}  \\ \hline

\end{tabular}}
\end{center}
\end{table}
\setlength{\tabcolsep}{1.4pt}

\subsection{Validation of the Different Distractors}


In this section, we explore the progressive integration of various distraction signals into the attention model, including gazing off-screen (via both gaze and head models), drowsiness, speaking, and unattended-screen distractors. Our exploration is conducted on two datasets: the real distraction dataset and the testing subset of the gazing dataset. We report the attention performance using the G-mean and F1 scores. Note that we do not evaluate the drowsiness and speaking signals on the gazing dataset, as it primarily focuses on gaze-related behavior.


Table~\ref{various_signals} shows the performance when adding the different distraction signals. We started with the most frequent distractor, gazing off-screen, and gradually added other distractors. Independently testing less frequent distractors, such as drowsiness and speaking, will yield low results due to their limited presence in the testing datasets. From the results, we can first conclude that the integration of all distraction signals contributes to enhanced attention detection. Second, the improvement in attention detection is consistent across both desktop and mobile devices. Third, the mobile sessions in the real dataset show significant head movements when gazing away, which are easily detected, leading to higher performance for mobile devices compared to desktops.
Fourth, adding the drowsiness signal has relatively slight improvement compared to other signals, as it's usually rare to happen. Finally, the unattended-screen signal has relatively larger improvement on mobile devices  compared to desktops, as mobile devices can be easily left unattended. 

Furthermore, our model is compared with a previously established work in ad testing, AFFDEX 1.0 \cite{mcduff2016affdex}. Comparing the performance of our attention model, or even solely the head model for gazing off-screen to the head model proposed in AFFDEX 1.0, shows a significant performance improvement across both desktop and mobile devices. This improvement is a result of incorporating head movements in both the yaw and pitch directions, as well as normalizing the head pose to account for minor changes. The pronounced head movements in the real mobile dataset have caused our head model to perform similarly to AFFDEX 1.0. Unfortunately, we were unable to compare our results with those from \cite{schulc2019automatic} due to the unavailability of their code.

\begin{figure*}[!t]
  \centering{\includegraphics[width=0.99\linewidth]{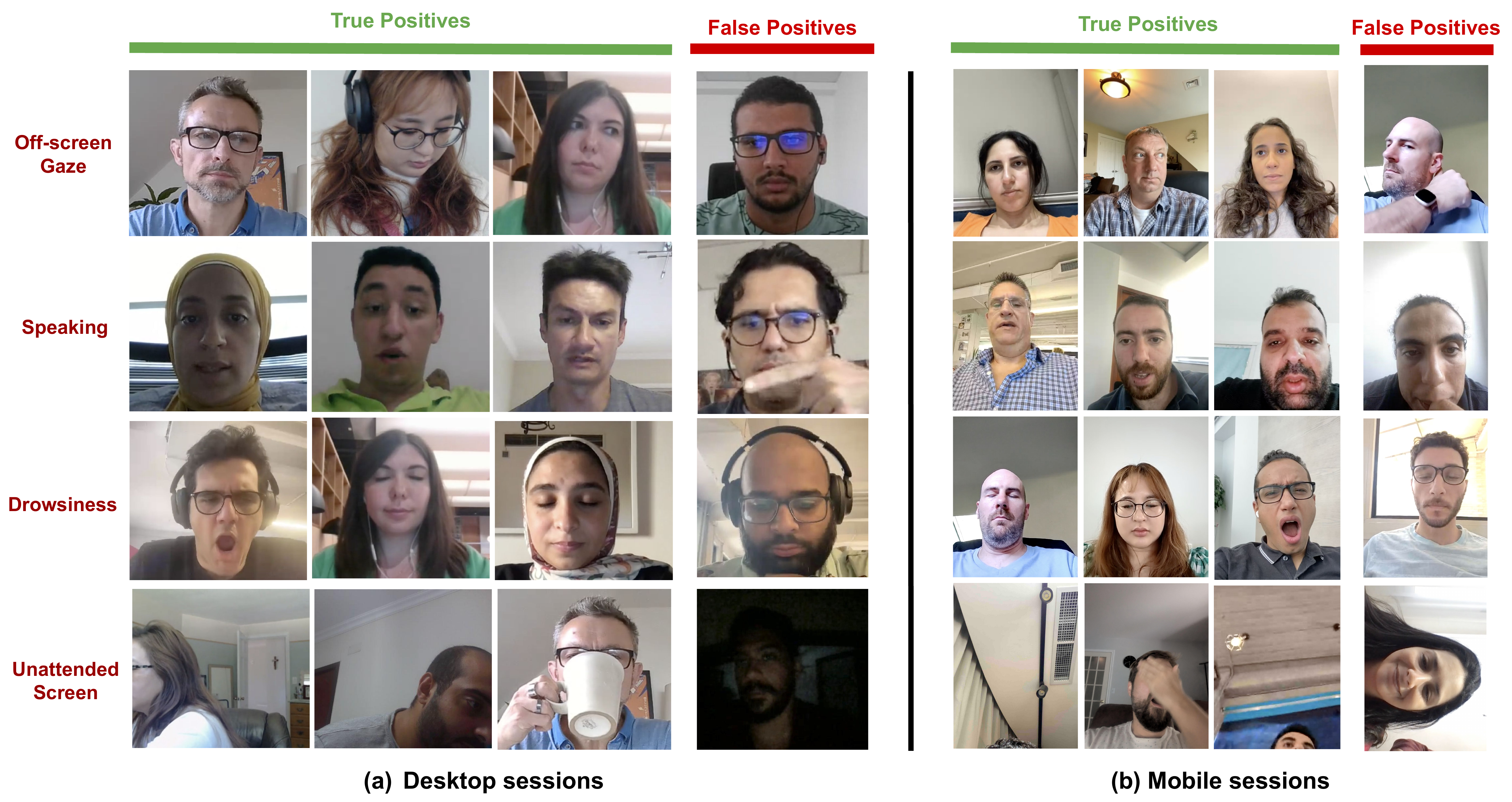}}
  \caption{Examples showing some true and false positives detected for the different distraction signals in our attention architecture. Examples are split depending on the device type (desktop and mobile).}
  \label{attention_results_frontal}
\end{figure*}

In Figure \ref{attention_results_frontal}, we present qualitative results illustrating the performance of our proposed attention model across desktop and mobile devices. Each row comprises a series of facial images representing both true and false positives for each distraction signal. The results indicate that our model effectively detects various distractors in uncontrolled settings. However, it may occasionally produce false positives in certain edge cases, such as severe head tilting while maintaining gaze on the screen, some mouth occlusions, excessively blurry eyes, or heavily darkened facial images. Figure \ref{attention_results_non_frontal} illustrates our model's ability to successfully identify various distractors in scenarios where participants use multiple screens.





\section{Conclusions and Future work}

Our work presents a novel architecture designed to monitor viewers attention during online ads. By integrating two facial analysis toolkits, AFFDEX 2.0 and SmartEye SDK, we extract low-level features including expressions, head pose, and gaze direction. These features are then used to derive high-level features such as fine-tuned gaze estimations, yawning, speaking activity, and more. This allows us to identify four key distractors: off-screen gaze, drowsiness, speaking, and unattended screens. Notably, our architecture customizes gaze settings based on the device type (desktop or mobile). We validate our approach using four datasets, each annotated for specific distractors. Results indicate promising performance in detecting attention across both desktop and mobile devices. In our future work, we aim to enhance our attention detection by refining our approach to determine if the gaze intersects with the ad being played, specifically when the ad is not played in the full-screen mode. Additionally, we plan to explore the detection of additional distractors such as eating or drinking.   

\bibliographystyle{splncs04}
\bibliography{main}
\end{document}